%% file: acl_latex.tex
\pdfoutput=1
\documentclass[11pt]{article}
\usepackage[table, dvipsnames]{xcolor}
\usepackage{emnlp}
\usepackage{times}
\usepackage{latexsym}
\usepackage{todonotes}
\usepackage{makecell}
\usepackage{booktabs}
\usepackage{array}
\usepackage[inline]{enumitem}
\usepackage{tcolorbox}
\usepackage{CJKutf8}

\usepackage[T1]{fontenc}
\usepackage[utf8]{inputenc}
\usepackage{CJKutf8}

\usepackage{microtype}

\usepackage{graphicx}
\usepackage{subcaption}
\usepackage{mwe}
\usepackage{hyperref}

\newcommand{\code}[1]{{\ttfamily#1}}

\title{DiQAD: A Benchmark Dataset for \\ End-to-End Open-domain Dialogue Assessment}

\author{
Yukun Zhao\textsuperscript{\rm 1,2}\thanks{~~Contributed equally.} \quad Lingyong Yan\textsuperscript{\rm 2}\footnotemark[1] \quad Weiwei Sun\textsuperscript{\rm 1} \quad Chong Meng\textsuperscript{\rm 2}
\\
\textbf{Shuaiqiang Wang}\textsuperscript{\rm 2} \quad \textbf{Zhicong Cheng}\textsuperscript{\rm 2} \quad \textbf{Zhaochun Ren}\textsuperscript{\rm 3}\thanks{~~Co-corresponding authors.} 
\quad \textbf{Dawei Yin}\textsuperscript{\rm 2}\footnotemark[2] \\ 
\textsuperscript{\rm 1}Shandong University, Qingdao, China \quad
\textsuperscript{\rm 2}Baidu Inc., Beijing, China \\ \textsuperscript{\rm 3}Leiden University, Leiden, The Netherlands\\
\texttt{\{zhaoyukun02,yanlingyong\}@baidu.com,sunnweiwei@gmail.com}\\
\texttt{\{mengchong01,wangshuaiqiang,chengzhicong01\}@baidu.com} \\
\texttt{z.ren@liacs.leidenuniv.nl,}~\texttt{yindawei@acm.org}
}

\begin{document}
\maketitle
\begin{abstract}
Dialogue assessment plays a critical role in the development of open-domain dialogue systems.
Existing work are uncapable of providing an end-to-end and human-epistemic assessment dataset, while they only provide sub-metrics like coherence or the dialogues are conversed between annotators far from real user settings.  
In this paper, we release a large-scale dialogue quality assessment dataset (DiQAD), for automatically assessing open-domain dialogue quality.
Specifically, we (1) establish the assessment criteria based on the dimensions conforming to human judgements on dialogue qualities, and (2) annotate large-scale dialogues that conversed between real users based on these annotation criteria, which contains around 100,000 dialogues.
We conduct several experiments and report the performances of the baselines as the benchmark on DiQAD.
The dataset is openly accessible at \url{https://github.com/yukunZhao/Dataset_Dialogue_quality_evaluation}.

%Many previous studies adopt user satisfaction or task success rate as the evaluation metrics, which is usually less capable of guiding dialogue systems to generate human-like and high-quality responses.
%In this paper, we release a large-scale Chinese dataset--DiQAD, for automatically assessing open-domain dialogue quality.

\end{abstract}

% Entries for the entire Anthology, followed by custom entries
\input{1-introduction}
\input{2-related_work}
\input{3-task_formulation}
\input{3_build_dataset}
\input{4-experiments}
\input{5-conclusion}
\newpage

\section*{Limitations}
The dialogue contents are written in Chinese, and hence the dataset can only be used to evaluate Chinese models. The impact may be restricted by its language. 
Besides, the benchmark models are absent from fine-tuning larger pre-trained models such as ChatGLM~\cite{du2022glm} and Vicuna~\cite{chiang2023vicuna}.
We cannot promise the current benchmark provide the strongest baseline. We believe that larger models and more fine-grained tuning would achieve better performance.
Finally, the quality annotation has been simplified to 0-2 scale for higher labeling accuracy. More degree of labels may help to differentiate the dialogue quality more elaborately. We will study the above limitations in our future work.

\section*{Ethics Statement}
We acknowledge the importance of the ACM code of Ethics and totally agree with it. 
We ensure that this work is compatible with the provided code, specifically in terms of providing the dialogue dataset. We use the dataset in accordance with copyright terms and under the licenses of its provider. 

\paragraph{Licensing}We collect the dialogues that have been licensed from both the users and the respondent. All the dialogues have been granted to publish by their owner, i.e., the WenYiWen platform. 
\paragraph{Personal Information} We collect dialogue texts without user information. The collecting procedure obeys the privacy policy according to local laws.
\paragraph{Toxic\&Privacy} All the dialogues have been examined by the WenYiWen audit team one by one, to avoid pornography, offensiveness, profanity, privacy, and toxicity risks.

\section*{Acknowledgements}
 
This work was supported by 
the National Key R\&D Program of China with grant No.2020YFB1406704, 
the Natural Science Foundation of China (62272274, 61902219, 61972234),
the Natural Science Foundation of Shandong Province (ZR2021QF129). 
We thank the WenYiWen platform for providing the original dialogues and helping us to filter privacy and toxic dialogues. 
%All content represents the opinion of the authors, which is not necessarily shared or endorsed by their respective employers and/or sponsors.

\bibliography{custom, anthology}
\bibliographystyle{acl_natbib}

\input{10_appendix}

\end{document}

%% file: 1-introduction.tex
\section{Introduction}
Open-domain dialogue system (ODS) is quite popular in artificial intelligence~\cite{serban_building_2016,huang_challenges_2020,bae_building_2022}, especially with the remarkable performance achieved in large language models (LLMs)~\cite{wei_2022_finetunedlanguagemodels, ouyang2022training, OpenAI2023GPT4TR}. 
Dialogue assessment is critical to the development of open-domain dialogue systems~\cite{deriu_survey_2020} as it guides what constitutes good conversational dialogues.
% Recently, as beneficiaries of the large language models (LLMs), e.g., FLAN~\cite{wei_2022_finetunedlanguagemodels}, and the GPT series~\cite{brown_2020_languagemodelsare,OpenAI2023GPT4TR}, open-domain dialogue systems have demonstrated their powerful capabilities in language understanding and utterance generation~\cite{2022DEAM}.

However, open-domain dialogue assessment still remains a challenging task~\cite{deriu_survey_2020, le2023improving}.
On the one hand, open-domain dialogue assessment is complicated and costly.
A straightforward solution is the human-based method, i.e., recruiting human evaluators to interact with the dialogue systems and provide their feedback~\cite{kelly2009questionnaires,dailydialog}, which is often expensive and less reproducible.
Other studies~\citep{onetomany2017, 2019Investigating, 2020Improving} release dialogues with references, which evaluate dialogue responses by comparing the generated responses with manually annotated references~\citep{yang2015wikiqa, rajpurkar2016squad, clark2016my, bleu2002}.
However, their performance is usually limited to reference coverage.
Reference-free evaluation~\cite{2022DEAM, learn-unref-based, 2020Improving,2020precise} uses crafted samples to train evaluation models due to the lack of annotations, which only capture coarse-grained metrics like coherence.

\begin{figure}[!t]
 \centering
 \includegraphics[width=\linewidth]{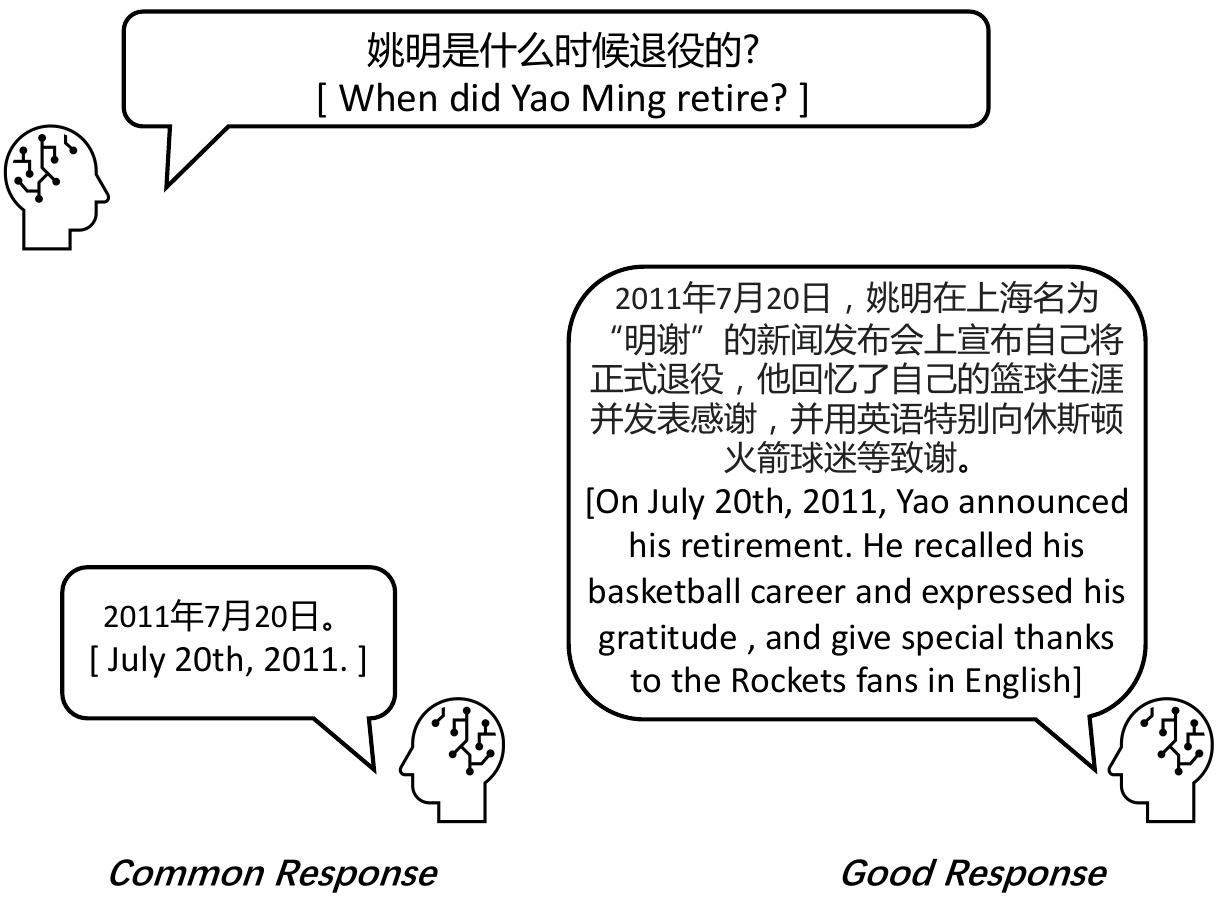}
 \caption{Both responses are satisfied, but the right one achieves higher quality due to its richer informativeness.}
 \label{fig:fig1}
\end{figure} 

On the other hand, most previous settings are less capable of end-to-end evaluation to determine whether the models perform like real humans.
For example, the assessment datasets~\cite{Mehri2020USRAU,gopalakrishnan2019topical} focus on evaluating particular aspects of dialogues, e.g., coherence or diversity, while ignoring the overall quality.
As shown in Figure~\ref{fig:fig1}, the previous settings consider both responses as satisfied ones and ignore their informativeness differences, lacking higher requirements like real humans.
Furthermore, previously released datasets~\cite{young2022fusing,komeili2021internet,gopalakrishnan2019topical,sun2021simulating} consist of dialogues conversed
between recruited annotators and bots or between annotators, which varies from practice.
As a consequence, the evaluating models learned from these datasets are far from optimal.

In this paper, we release a large-scale dataset - DiQAD (\textbf{Di}alogue \textbf{Q}uality \textbf{A}ssessment \textbf{D}ataset) for open-domain dialogue quality assessment, which aims to learn an atomically evaluating model to guide the dialogue system in producing both satisfied and high-quality responses.
Specifically, we first establish unified human-epistemic quality criteria for open-domain dialogues based on 6 quality dimensions: grammaticality, relevance, consistency, empathy, proactivity, and informativeness~\cite{finch2020towards}.
We set the quality label as a 3-scale holistic score (from 0 to 2) by considering the above dimensions, identifying higher quality dialogues that provide more comprehensive information and actively engage in the conversations using a score of 2.
After that, we collect a substantial number of real dialogues from an online conversation platform and hire experienced annotators to annotate them following the above quality criteria.
In total, \textbf{DiQAD} contains around $100$k dialogues with around 0.8 million dialogue utterances across 6 common domains.

On the DiQAD, we conduct extensive experiments to verify the performance of different models. 
Additionally, the cross-domain experiments on DiQAD show that the best model learned on our dataset can generalize to unseen domain dialogues.

The contributions of this paper are twofold:
\begin{itemize}
    \item This paper defines the rules for assessing dialogue quality and constructs a large-scale dialogue evaluation dataset. To the best of our knowledge, this is the first large-scale dataset focused on dialogue quality assessment.
    \item This paper conducts benchmark experiments on DiQAD and finds that recent large language models (e.g., ChatGPT, ChatGLM) are less capable of discriminating high-quality dialogues. % lists the performances of the most recently proposed dialogue evaluating models.
\end{itemize}

%% file: 2-related_work.tex
\section{Related Work}
\paragraph{Dialogue System} 
Dialogue systems are quite popular, with the remarkable performance of large language models~\cite{ouyang2022training, OpenAI2023GPT4TR}, and they are usually classified into task-oriented and open-domain dialogue systems~\cite{ni2022recent}. 
The task-oriented dialogues aim to solve specific tasks in a certain domain, which are supported by pipeline systems~\cite{li2017end,cheng2020conversational,williams2017hybrid,golovanov2019large} or end-to-end systems~\cite{ni2022recent,le2020uniconv, wang2019incremental, he2020amalgamating}. 
The open-domain dialogue system consists of chat-oriented systems aiming to converse with users without task and domain restrictions~\cite{tao2019one,feng2020regularizing,xu2020conversational, song2020generate,miao2019cgmh,wu2021controllable}, and conversational question-answering systems which are developed to answer specific questions~\cite{liu2019roberta,sun2019ernie,qu2019attentive}.

%The chat-oriented dialogues . Many work propose to handle the challenges including context awareness~\cite{tao2019one,feng2020regularizing,qiu2020structured}, response coherence~\cite{liu2019vocabulary, xu2020conversational}, response diversity~\cite{li2015diversity,vijayakumar2016diverse,qiu2019training}, personality~\cite{liu2020you, kim2020will, song2020generate}, controllable generation~\cite{hu2019parabank,miao2019cgmh,wu2021controllable}. 
%The QA dialogues are developed to answer specific questions while the questions are open domain, which is a special form of task-oriented dialogues, including single-turn QA~\cite{devlin2018bert,lan2019albert,liu2019roberta,sun2019ernie} and multi-turn QA~\cite{qu2019attentive}. 
%To evaluate the mentioned dialogues, there are several methods focused on different aspects.

\paragraph{Dialogue Evaluation}
There are two directions for dialogue evaluation: human evaluation and automatic evaluation.
The human evaluation involves recruiting experts to test a dialogue system and then collecting questionnaires (i.e., user ratings)~\cite{kelly2009questionnaires,ashwin2017conversational,dailydialog}. 
For the dialogues with predefined tasks, metrics like task-completion rate and task-completion cost are calculated from user interactions~\cite{walker1997paradise,bodigutla2019multi} or user simulators~\cite{schatzmann2007agenda,2009agenda,zhang2020evaluating,sun2022metaphorical} to evaluate the dialogues.

%The real users' interactions make the evaluation high-cost and hard to reproduce~\cite{balog2021conversational}. For the dialogues whose tasks are specific, user simulators are developed to provide simulated user dialogues and the automatic metrics are calculated(e.g., the task-completion rate and task-completion efficiency) for the evaluation~\cite{schatzmann2007agenda,2009agenda,zhang2020evaluating,sun2022metaphorical}. This method is not applicable to open domain dialogues when they have no clear tasks. 

Another way for automatic evaluation is to evaluate the quality of dialogue contents
% , including reference-based and reference-free metrics. %Another way is to evaluate the dialogue content by evaluating the response's correctness and appropriateness. % (e.g., coherence, diversity, and informativeness).
One is reference-based evaluation, which measures the similarity between the generated responses and the ground-truth one, including correctness metrics such as MAP, MRR, EM, F1, and accuracy~\cite{yang2015wikiqa, rajpurkar2016squad, clark2016my}. Word-overlap statistics like BLEU, ROUGE, and METEOR~\cite{bleu2002, rouge2002, METEOR}, as well as neural metrics~\cite{Sato2020, RUBER, zhao2023evaluating}, are calculated. However, it may suffer from the one-to-many problems~\cite{onetomany2017}, even with multi-references~\cite{2019Investigating, 2020Improving}. %Then the subsequent work comes to the reference-free metrics. %~\cite{ADEM2017, liang2020beyond, Re-ADEM2019},
Besides, recently released datasets are dialogues that occur between recruited annotators and bots~\cite{young2022fusing}, or between annotators themselves~\cite{wang2021naturalconv, smith2020can, komeili2021internet, gopalakrishnan2019topical}, or happen in certain scenarios like English practice~\cite{dailydialog}. The conversational topics and contents are handcrafted and limited compared to real users’ settings.

The other one is reference-free evaluation, which train a classification model using users' ratings~\cite{liang2020beyond} or crafted samples~\cite{2022DEAM, learn-unref-based, 2020Improving,2020precise} for coarse-grained evaluation like coherence due to the short of annotated labels.
%, while the user ratings are noisy, unreliable and impracticable~\cite{evanini2008caller,schmitt2015interaction}.
%The others try to use crafted samples~\cite{2022DEAM, learn-unref-based, 2020Improving,2020precise} for coarse-grained evaluation such as coherence. 
Recent work~\cite{Mehri2020USRAU,deriu2020spot_bestp,Know-More-19,sun2021simulating,le2023improving} use annotated samples to train the evaluation model.
%while the expert judgment uses unified criteria alleviating the subjective and is shown to be more practicable~\cite{ultes2013quality}. Then the expert annotations are more commonly used as a replacement for user ratings. 
The annotations for open-domain dialogues are desirable, but the previous work only focuses on sub-type of dialogues such as task-oriented dialogues~\cite{sun2021simulating}. For open-domain dialogue annotation, they~\cite{sun2021simulating,bodigutla2019multi,Mehri2020USRAU,gopalakrishnan2019topical} evaluate some sub-metrics like coherence, diversity, empathetic or factuality, lacking of compound and human-epistemic high-quality evaluation and the data size is limited.

%Compared with agenda-based user simulation\cite{schatzmann2007agenda,2009agenda} which only feed dialogue systems with the predefined user goal, \citep{zhang2020evaluating} present an approach for ensuring the consistency of response and use both automatic, and manual evaluation to compare existing conversational systems. \citep{sun2022metaphorical} propose a metaphorical user simulator that use historical strategies to improve the realism and transferability of the simulator. 

%For chit-chat dialogues, the evaluation mainly aims to assess the response appropriateness (e.g., coherence, relevance, and informativeness) or human likeness. One is the reference-based metrics which calculate the similarity between the generated responses and the ground truth such as word-overlap statistics\cite{bleu2002,rouge2002,METEOR} and neural methods\cite{Sato2020}. It is not applicable due to the difficulty to obtain ground-truth responses and one-to-many problems\cite{onetomany2017} even with multi-references\cite{2019Investigating, 2020Improving}. Then the subsequent work comes to the reference-free metrics\cite{ADEM2017, liang2020beyond, Re-ADEM2019}, which are trained using real user ratings as label\cite{liang2020beyond}, or crafted samples\cite{2022DEAM, learn-unref-based, 2020Improving}, or expert annotation samples modeling users' satisfaction on sub metrics including coherence, fluency, informative\cite{2022DEAM,deriu2020spot_bestp,Know-More-19}.
%mimicking human beings to generate fluent, coherent utterances and informative }. 

%% file: 3-task_formulation.tex
\begin{table*}[ht]
\centering\small
% \resizebox{\textwidth}{!}{
\begin{tabular}{@{}p{11.2cm}<{\centering} c p{2.8cm}<{\centering}@{}}
\toprule
%\hline
Dialogue & Score & Explanation \\
\midrule
\makecell[l]{\textbf{Q}: Any suggestions for buying iPhone 14 or 14 pro? \\
\textbf{R}: The iPhone is designed and marketed by Apple Inc. Apple is an American \\ multinational technology company ...} &  0 & \makecell[c]{Not relevant \\ between buying \\ and market}
\\
\midrule
\makecell[l]{\textbf{Q}: My girlfriend is annoyed at my ex. 
\\ 
\textbf{R}: Perhaps you remain affiliated with your ex. When you break up with your ex compl-\\etely, she would not be upset anymore.}
 &  1 
 & \makecell[c]{Give only one reason \\that do not provide \\ complete information}
\\
\midrule
\makecell[l]{\textbf{Q}: Which products are manufactured by Foxconn? \\
\textbf{R}: Foxconn manufactures electronic products including the iPhone, MacBook, Dell, HP \\ computers, and other consumer electronics.  
\\ \textbf{R}: Foxconn manufactures electronic products for major American, Canadian, Chinese, \\Finnish, and Japanese companies. It is a multinational electronics contract manufacturer ...}
 &  2 & \makecell[c]{Provide the all \\desired information. } \\
\bottomrule
\end{tabular}
% }
\caption{Dialogue quality assessment examples following our criteria. The Q and R denote the question and response respectively. All contents are translated from Chinese.}
\label{tab:annotationsample}
%\vspace{-0.2cm}
\end{table*}

\section{Open-domain Dialogue Quality Assessment}
\label{sec:task_formulation}

% We define the open-domain dialogue quality assessment assigning a unified score $s$ to a given dialogue $D=\{q_1, r_1, ..., q_i, r_i, ...\}$, where $q_i, r_i$ are the utterances of the questioner and respondent. 

This paper treats open-domain dialogue quality assessment as a critical task, which differentiates higher quality dialogues based on real users' dialogues, to guide the future dialogue generation towards more satisfying and human-epistemic responses.

To this end, the quality assessment criteria are required to reflect higher quality for human cognition as much as possible.
The human demands upon dialogue quality are usually regarded as entailing several fine-grained dimensions~\cite{deriu_survey_2020,smith_human_2022}.
Inspired by a fully analyzed human-epistemic evaluation~\cite{finch2020towards}, we adopt the following 6 dimensions to establish our quality assessment criteria (see examples in Table~\ref{tab:annotationsample}):
\begin{itemize}
\setlength{\topsep}{1pt}
\setlength{\itemsep}{2pt}
\setlength{\parskip}{1pt}
    \item \textbf{Grammaticality}: Whether the utterances are fluent, readable, and free of grammatical and semantic errors.
    \item \textbf{Relevance}: Whether the responses logically match and are coherent with the questions.
    \item \textbf{Consistency}: Whether the utterances provide a consistent persona and no contradictions with the previously provided utterances.
    %\item Engagingness: whether the response gives interesting content.
    \item \textbf{Empathy}: Whether the respondent comprehends the user's feelings and appropriately reacts to emotional expressions (e.g., appeasing). %The metric can be neglected in non-emotional dialogues.
    \item \textbf{Proactivity}: Whether the respondent responds actively, provides useful extensions, and moves the dialogue to new topics.
    %\item Completeness: whether the response provides all the complete related information.
    \item \textbf{Informativeness}: Whether the responses provide useful, specific, and sufficient information.
\end{itemize}

Based on the above dimensions, we conduct a 3-scale (0, 1, 2) quality assessment.
A dialogue is set to at least answer the user's question; thus, the quality is at least a binary category of 1 and 0 to distinguish whether the user's question is resolved or not. Furthermore, we aim to identify high-quality dialogues that guide future dialogue generation towards higher quality, like humans. To achieve this, we introduce an additional category of 2 indicating a high-quality rating.\footnote{Our annotators show lower annotation agreement under a more fine-grained, like 1-5 scale assessment for dialogue quality. We leave the 1-5 scale version for future work.}

The quality assessment includes turn-level and dialogue-level evaluation. We provide dialogue-level quality evaluation criteria. Specifically, we assess the quality of dialogue according to the following instructions (The examples are shown in Table~\ref{tab:annotationsample}, and detailed instructions and more examples are shown in Appendix~\ref{subsec:annotation_instruction}):
\begin{itemize}
\setlength{\topsep}{1pt}
\setlength{\itemsep}{2pt}
\setlength{\parskip}{1pt}
    \item \textbf{Low quality (score 0)}: The dialogue fails to meet the basic requirements, including grammaticality, relevance, consistency, and empathy. As shown in Table~\ref{tab:annotationsample}, the response of the first case is irrelevant to its question. Thus we label it 0-class directly without considering other metrics.
    \item \textbf{Moderate quality (score 1)}: The dialogue can basically answer the user's question, providing grammatical, relevant, consistent, and empathetic responses but fails to meet proactivity and informativeness. As the second case in Table~\ref{tab:annotationsample}, the response is only one reason that fails to supply sufficient information to meet the informative requirement.
    \item \textbf{High quality (score 2)}: The dialogue meets all the dimensions mentioned above. It is a higher requirement in addition to correctly answering questions on 1-class dialogue, which provides more comprehensive information and actively engages in conversations with the user, as shown in the third case in Table~\ref{tab:annotationsample} and Figure~\ref{fig:guide_annotation_2} in Appendix~\ref{subsec:annotation_instruction}. % the response provides complete and useful extensions describing the markets of the products besides the basic requirements.
\end{itemize}

In the next section, we introduce how we build the DiQAD based on the above criteria.

%% file: 3_build_dataset.tex
\section{DiQAD: A Large-scale Dataset for Dialogue Quality Assessment }
\label{sec:build_dataset}
%%data_construction 数据数据集如何采集，如何标注
%To guide the automatic model to learn how to evaluate the open-domain dialogue quality like human beings, we build a large-scale dialogue evaluation dataset called DiQAD.

\noindent In this section, we describe how we construct the DiQAD based on the above annotation instructions in detail.
We first collect amounts of real dialogues from an online platform. 
Then, we recruit experienced annotators to annotate these dialogues.
In the last part of this section, we will show the detailed statistics of DiQAD.

% We propose to build a dataset for the conversational and task-oriented dialogue fusing evaluation. Due to the lack of high-quality real human-written dialogues, we sample dialogues from real commercial conversational agents and annotate hundreds of thousands of data for the community. The subsection~\ref{sec:data_pre} describes the data preparation. We describe how to annotate the data in subsection~\ref{sec:annotation}.

\subsection{Real Dialogue Collection}
\label{sec:data_pre}
In the first step, we collect real dialogues in text from WenYiWen\footnote{\url{https://wen.baidu.com}}.
WenYiWen provides online conversational services, in which users can chat with others and obtain required information from experienced experts.
Users can propose their questions to the platform, and then experienced experts can browse them and choose their interested ones to start a dialogue.
After that, the dialogue takes place between the user and the expert until the user obtains their satisfied information or the duration reaches its limit.
WenYiWen provides domain and topic tags for user questions to help experts select their related questions. The experts are invited from various professions, such as lawyers, university professors, college students, and car mechanics.
Besides, to attract enough experts, users are asked to pay for their questions, and the experts can receive cash rewards based on the dialogue quality.

Specifically, we collect real dialogues in 6 domains from publicly accessible WenYiWen dialogues set\footnote{For each WenYiWen dialogue, users are asked whether to open their dialogues to the public. If users agree to open, then the dialogue will become publicly accessible.}.
The domains we selected consist of emotion, digital, legal, education, car, and finance dialogues.
%The statistics are shown in Table~\ref{table:sentiment}, including the number of dialogues and the number of utterances per dialogue.
%Additionally, we filter out the samples without a response and reduce the proportion of dialogues containing one turn.
Consequently, the collected dialogues for annotation are more than 100 thousand (The statistics of collected dialogues are shown in Section~\ref{sec:statistics}).

 \begin{figure*}[ht]
\centering
    \begin{subfigure}[b]{0.425\textwidth}
    \centering
    \includegraphics[width=\textwidth]{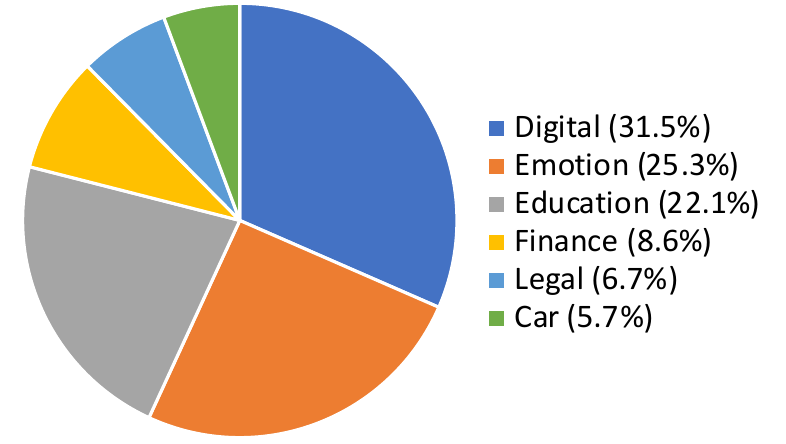}
    \caption{}
    \label{fig:dataset_distribution_a}
    \end{subfigure}
\hfill
    \begin{subfigure}[b]{0.554\textwidth}
    \centering
    \includegraphics[width=\textwidth]{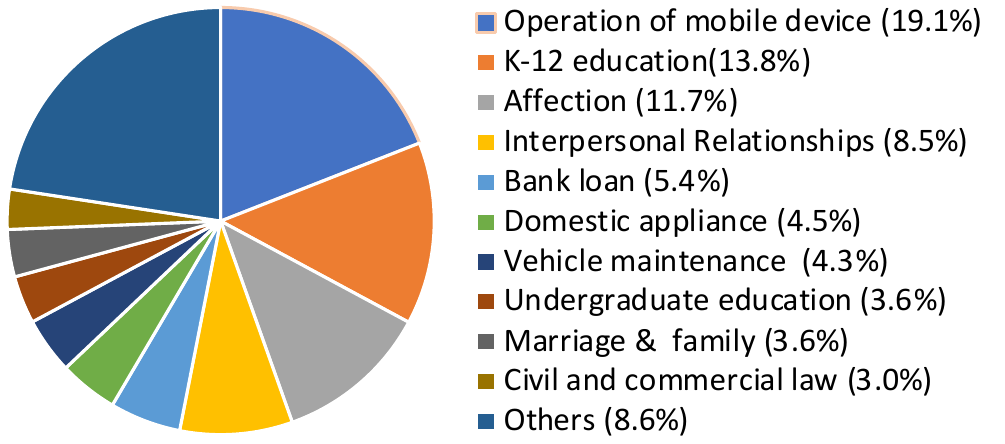}
    \caption{}
    \label{fig:dataset_distribution_b}
    \end{subfigure}
\caption[]
{The domain (a) and topic (b) distributions of DiQAD.} 
\label{fig:dataset_distribution}
\end{figure*}

\subsection{Quality Annotation}
\label{sec:annotation}
Next, we recruit dozens of experienced annotators to annotate the dialogue quality on a crowd-sourcing platform developed by ourselves. %The details about annotators are shown in Appendix~\ref{sec:annotators}.

The annotation process is an annotation-checking procedure, or repeated annotation-checking procedures if the annotation accuracy of the first round of annotation does not match the requirements. The dialogues are divided into multiple batches for annotation, with each batch consisting of 2000 samples. The annotators are divided into several groups, with each group annotating the batches of data. 

For each dialogue, the quality label has been agreed upon and confirmed by 2 annotators. The annotator is first presented with the whole dialogue text. To guarantee the labeling reliability, the reading duration of each dialogue is required to be no less than a fixed time.
After that, annotators are asked to label the quality score following the annotation criteria in section~\ref{sec:task_formulation} and the detailed instructions in Appendix~\ref{subsec:annotation_instruction}.
In addition to the quality labels, annotators are also asked to give explicit reasons why the labels are decided to improve annotation quality.

After annotating a batch of data, we have a professional quality controller to check the label accuracy. (The roles and distribution of crowd-source annotators can be found in Appendix~\ref{sec:annotators}). Our acceptance criterion is to achieve an annotation accuracy of 92\% or above. More detailed information about annotation is shown in Appendix~\ref{sec:appendix}.

% Specifically, the annotators are divided into $N$ groups, the collected dialogues are randomly split into $M$ groups ($N \gg M$).
% Each dialogue group is assigned to one annotator group.
% And each dialogue is randomly assigned to at least two annotators in the same annotator group.
% If two annotators give the same quality label to a dialogue, this label will be annotated as the final label of this dialogue.
% Otherwise, the dialogue will be reviewed by the third annotator to solve the disagreement between the first two annotators.
% Besides, we calculate the agreement ratio of each dialogue group after the dialogue group is fully annotated.
% And if the agreement ratio is lower than 90\%, this dialogue group will be re-annotated by another group.

%Finally, we invite other well-educated annotators to sample and check each batch of data ensuring the annotation accuracy is higher than 92\%. 

%%标准、多人拟合、额外问题

\subsection{Dataset Statistics}
\label{sec:statistics}

%% 1. 与已有数据集的统计对比（表格）
%% 2. 数据集各领域的统计对比（图）
%% 3. 与人的反馈的分析对比
Finally, we obtain DiQAD, consisting of 100k dialogues across 6 different domains. Its basic statistics are shown in Table~\ref{table:statistics_compare}. DiQAD contains over 100k dialogues, where each dialogue consists of around 8 utterances and 334 tokens on average. In Table~\ref{table:statistics_compare}, we also compare our proposed DiQAD with recent popular open-domain datasets, DailyDialogue~\citep{dailydialog}, BlendedSkillTalk~\citep{smith2020can}, NaturalConv~\citep{wang2021naturalconv}, FusedChat~\citep{young2022fusing}, WizInt~\citep{komeili2021internet}, USR~\citep{Mehri2020USRAU}, FED-Dial~\citep{mehri2020unsupervised}, and TopicalChat~\cite{gopalakrishnan2019topical}, while the first 5 datasets have no quality annotation. NaturalConv and DiQAD are written in Chinese, and the others are in English. The scale of DiQAD is larger, and it is a compound human-epistemic evaluation in real users' dialogues compared with all other datasets.

% \begin{table*}[!t]
% \centering
% % \resizebox{\textwidth}{!}{
% \begin{tabular}{@{}llllllll@{}}
% \toprule
% %\hline
%  & All & Emotion & Digital & Legal & Education & Car & Finance \\
% \midrule
% Dialogues   &  100k & 27k & 34k & 7k & 24k & 6k & 9k \\
% % \midrule
% Turns Per Dialogue  & 5.5 & 6.1 & 4.6 & 6.2 & 5.7 & 4.9 & 5.5\\
% % \midrule
% Questions Per Dialogue  & 3.4 & 3.6 & 2.9 & 3.9 & 3.6 & 3.0 & 3.3\\
% % \midrule
% Responses Per Dialogue  & 4.5 & 5.8 & 3.2 & 5.3 & 5.0 & 3.2 & 4.7 \\
% % \midrule
% Tokens Per Dialogue  & 334 & 377 & 325 & 438 & 297 & 278 & 291\\
% % \midrule
% % Words Per Question  & 22.3 & 34.6 & 16.4 & 22.7 & 17.5 & 15.9 & 18.4 \\
% % \midrule
% % Words Per Response  & 57.3 & 43.1 & 85.9 & 65.8 & 47.3 & 73.0 &	49.5 \\
% %\hline
% \bottomrule
% \end{tabular}
% % }
% \caption{Basic statistics of DiQAD. Each turn may contain several questions or responses.}
% \label{tab:statistics_dataset}
% %\vspace{-0.2cm}
% \end{table*}

\begin{table}[!t]
\centering
\setlength\tabcolsep{1.5pt}
\begin{tabular}{@{}lcccc@{}}
\toprule 
Dataset &   \#insts.\ & \#uttrs.\ & \#tokens & q-label \\
\midrule
%\midrule
DailyDialogue & 13k	& 7.9	& 114.7 & N \\
BlendedSkillTalk & 5k	& 11.3	& 155.4 & N \\
%KDConv & 5k & 19 & 377\\ %domain=3
NaturalConv & 19k & 20.1 & 244.8 & N\\
FusedChat & 10k	& 5.8	& 65.2 & N \\
WizInt & 10k	& 9.7	& 185.2 & N \\
\midrule
USR & 660	& 9.3	& 180.2  & Y\\
FED-Dial & 125	& 12.7 & 113.8 & Y \\
TopicalChat & 10k & 21	& 399 & Y \\
\midrule
DiQAD & 100k	& 7.9	& 334.1 & Y \\
\bottomrule
\end{tabular}
\caption{The statistics of DiQAD and other open-domain dialogue evaluation datasets. We list the number of dialogue instances, the mean number of utterances per dialogue,  the mean number of tokens per dialogue, and whether containing quality labels.}\label{table:statistics_compare}
\vspace{-0.6cm}
\end{table}

The domain distributions of DiQAD are shown in Figure~\ref{fig:dataset_distribution_a}.
We see that the top 3 domains in DiQAD are digital, emotion, and education, with the smaller ones are finance, legal, and car.
We utilize the built-in topic tags mentioned in section~\ref{sec:data_pre} and show the top 10 topics in DiQAD in Figure~\ref{fig:dataset_distribution_b}.
We see that DiQAD spreads over diverse topics such as the operation of mobile devices, affection, bank loans, etc., which can reflect the real users' settings to some extent.
%We see that dialogues on relationships focusing on chat need more turns around 6.1, and the questions and responses are more accordingly. The dialogues on digital focusing on information seeking, are more like QA dialogues in that the responses are relatively longer. 

%% file: 4-experiments.tex
\section{Experiments}
In this section, we conduct and illustrate benchmark experiments on DiQAD, including examining the overall performance of benchmark models (\S\ref{sec:backbone}), conducting a detailed analysis based on the best benchmark model (\S\ref{sec:detailed_analysis}), comparing to other reference-free evaluation methods (\S\ref{sec:reference-free}), and performing hyperparameter analysis (\S\ref{sec:hyperparameter}).

\subsection{Experimental Setup}
\paragraph{Dataset.}
We randomly split DiQAD into training, validation, and test sets, where the validation and test sets consist of 10,000 dialogues, and the remaining dialogues are used for training. For all models trained on the training set, we tune their parameters on the validation set and report the performance results on the test set. The final test set contains 1,958 samples classified as score 0, 3,628 samples classified as score 1, and 4,414 samples classified as score 0.

% We randomly split the DiQAD into training/validation/test sets with 87,596/10,000/10,000 dialogues. The ratio of 2-score, 1-score, 0-score are 43\%, 36\%, 21\% in the training set, 44\%, 36\% and 20\% in the validation set, 44\%, 35\% and 21\% in test set. We tune the parameters of the models on the validation set and report the performance on the test set.

\paragraph{Metrics.}
We employ the following metrics of baseline models\footnote{We use the implementation from SciPy (\url{https://scipy.org/}) to calculate the Spearman and Pearson scores.}:
(i) \emph{Acc}, the accuracy of the model predictions;
(ii) \emph{Unweighted average recall (UAR)}, the arithmetic average of
class-wise recalls;
(iii) \emph{Cohen's Kappa coefficient (Kappa)}~\citep{Cohen1960ACO}, which we used to measure the agreement between model-predicted labels and human-annotated labels;
(iv) \emph{Spearman's rank correlation coefficient (Spearman)}, the non-parametric measure of the correlation between model predictions and human labels;
(v) \emph{Pearson correlation coefficient (Pearson)}, the measure of linear correlation between model predictions and human labels.
% (vi) \emph{Bi-F1}, the F1-score for the dissatisfied (i.e., rating = $0$) class as the binary classification metric, which aims to evaluate whether the model can identify dissatisfied dialogue;
Besides, we also report the precision, recall, and F1-score of model predictions.

% \begin{table*}[!t]\centering
% \setlength\tabcolsep{5pt}
%  \resizebox{\textwidth}{!} 
% {
% \begin{tabular}{@{}l cccccccccc@{}}
% \toprule
% & Acc & UAR & Kappa & Rho & Spearman & Pearson & Bi-F1 & Precision & Recall & F1  \\
% \midrule
% \multicolumn{4}{@{}l}{\emph{ML-based methods}}\\
% Bayes & 52.67	& 28.53	& 24.03	& 55.84	& 22.64 & 24.52 & 16.61 & 48.35 & 47.55 & 47.56 \\
% LR & 60.98	& 32.99	& 36.83	& 64.43	& 37.66 & 39.60 & 26.82 & 57.74 & 54.98 & 55.21 \\
% XGBoost &63.37	& 34.56	& 40.54	& 66.31	& 42.21 & 43.35 & 36.55 & 62.06 & 57.59 & 58.21 \\
% \midrule
% \multicolumn{4}{@{}l}{\emph{NN-based methods}}\\
% GRU & 61.34	& 33.72	& 37.68	& 66.27	& 39.67 & 41.15 & 33.79 & 60.02 & 56.11 & 56.53 \\
% Transformer & 64.16	& 35.02	& 41.61	& 70.26	& 45.51 & 46.66 & 40.91 & 63.12 & 58.62 & 59.28\\
% BERT & 68.23	& 38.76	& 48.91	& 74.88	& 55.29 & 55.37 & 57.94 & 67.92 & 64.61 & 65.71 \\
% MENGZI & 69.25	& 39.79	& 50.91	& \textbf{74.96} & 56.28 & 56.29 & 59.82 & 68.28 & 66.31 & 67.07 \\
% ERNIE & \textbf{69.94}	& \textbf{40.96}	& \textbf{52.55}	& 74.00	& \textbf{57.39} & \textbf{57.25} & \textbf{62.96} & \textbf{68.68} & \textbf{68.27} & \textbf{68.46}\\
% \bottomrule
% \end{tabular}
% }
% \caption{Compare different backbone models on DiQAD. The boldface indicates the best result in terms of the corresponding metric.}\label{table:model-selection}
% \end{table*}

\begin{table*}[!t]\centering
\small
\begin{tabular}{lp{3.5em}<{\centering}p{3.5em}<{\centering}p{3.5em}<{\centering}p{3.5em}<{\centering}p{3.5em}<{\centering}p{3.5em}<{\centering}p{3.5em}<{\centering}p{3.5em}<{\centering}}
\toprule
& Acc & UAR & Kappa & Spearman & Pearson & Precision & Recall & F1  \\
\midrule
\multicolumn{4}{@{}l}{\emph{Classical methods}}\\
Bayes & 52.67	& 28.53	& 24.03	& 22.64 & 24.52 & 48.35 & 47.55 & 47.56 \\
LR & 60.98	& 32.99	& 36.83	& 37.66 & 39.60 & 57.74 & 54.98 & 55.21 \\
XGBoost &63.37	& 34.56	& 40.54	& 42.22 & 43.35 & 62.06 & 57.59 & 58.21 \\
GRU & 61.34	& 33.72	& 37.68	& 39.98 & 41.15 & 60.02 & 56.11 & 56.53 \\
\midrule
\multicolumn{4}{@{}l}{\emph{Transformer-based methods}}\\
Transformer & 64.16	& 35.02	& 41.61	& 45.52 & 46.66 & 63.12 & 58.62 & 59.28\\
BERT & 68.23	& 38.76	& 48.91	& 55.59 & 55.37 & 67.92 & 64.61 & 65.71 \\
MENGZI & 69.25	& 39.79	& 50.91	& 56.28 & 56.29 & 68.28 & 66.31 & 67.07 \\
ERNIE & \textbf{69.94}	& \textbf{40.96}	& \textbf{52.55}	& \textbf{57.39} & \textbf{57.25} & \textbf{68.68} & \textbf{68.27} & \textbf{68.46}\\
\midrule
\multicolumn{4}{@{}l}{\emph{LLM-based methods}}\\
ChatGLM-6B & 36.70 & 21.08 & 3.05 &  9.44 & 10.09 & 36.22 & 36.70 & 34.22
\\
% Vicuna-7B \\
ChatGPT & 39.30 & 21.88 & 4.70 & 12.83 & 12.51 & 37.61 & 36.47 & 36.28\\
ChatGPT + CoT & 41.00 & 24.91 & 10.68 & 17.79 & 17.33 &	30.77 & 31.14 & 30.38 \\

\bottomrule
\end{tabular}
\caption{Comparison of benchmark models on DiQAD. The best results of each metric are shown in bold.}\label{table:model-selection}
\end{table*}

\paragraph{Benchmark Models.}
We use the following two types of models as our benchmark models: classical methods, Transformer-based methods, and LLM-based methods.
\begin{itemize}
    \item \emph{Classical methods}: Naive Bayes (Bayes), Logistic Regression (LR), XGBoost, and GRU.

    \item \emph{Transformer-based methods}: Vanilla Transformer~\citep{Vaswani2017AttentionIA}, a 4-layer Transformer; BERT~\citep{devlin2018bert}; MENGZI~\citep{Zhang2021MengziTL}, an efficiently pre-trained Transformer using a 300G Chinese corpus; ERNIE~\citep{Sun2021ERNIE3L}, a 12-layer Transformer encoder pre-trained on a 4TB corpus consisting.

    \item \emph{LLM-based methods}: ChatGLM-6B~\cite{du2022glm}, an open-source Chinese LLM; and ChatGPT~\cite{ouyang2022training}, an LLM developed by OpenAI and accessed through \code{gpt-3.5-turbo-0301} API.
    
\end{itemize}

% \begin{itemize*}
%     \item \textbf{Naive Bayes (Bayes)}, the probabilistic classifier based on Bayes' theorem;
%     \item \textbf{Logistic Regression (LR)}, a linear classification model; 
%     \item \textbf{XGBoost}, a classification model based on gradient boosted trees algorithm; and
%     \item \textbf{GRU}~\citep{Cho2014OnTP}, a one-layer gated recurrent unit model with the hidden size 256.
% \end{itemize*}
% The \emph{Transformer-based methods} includes:
% \begin{itemize*}
%     \item \textbf{Vanilla Transformer}~\citep{Vaswani2017AttentionIA}, a 4-layer Transformer with random initialization;
%     \item \textbf{BERT}~\citep{devlin2018bert}, the popular Transformer-based language model;
%     \item \textbf{MENGZI}~\citep{Zhang2021MengziTL}, an efficiently pre-trained Transformer using 300G Chinese corpus;
%     \item \textbf{ERNIE}~\citep{Sun2021ERNIE3L}, a 12-layer Transformer encoder pre-trained on a 4TB corpus consisting of plain texts and a large-scale knowledge graph.
% \end{itemize*}

For Transformer-based models, we use the same data format as the input. Specifically, given a dialogue, we use ``\code{question:}'' and ``\code{response:}'' to annotate the sentences from the questioner and respondent, respectively. We then splice the dialogue into a long text and feed it into the models to get predictions. For Transformer-based models, we add two special tokens, \code{[CLS]} and \code{[SEP]}, to the beginning and end positions after tokenization. After encoding the dialogue using Transformer-based models, the embedding of the \code{[CLS]} token is fed into a 3-class classification layer to get predictions. For LLM-based methods, we use the instructions detailed in Appendix~\ref{sec:prompt} for few-shot learning.

\paragraph{Implementation.}
% We use the ERNIE 3.0 models publicly available at \url{https://github.com/nghuyong/ERNIE-Pytorch} in our experiments. 
The neural models are optimized using the cross-entropy loss. We set the batch size to $16$, the learning rate to $1e{-}5$, use the AdamW optimizer to optimize parameters, employ gradient clipping with a maximum gradient norm of $1.0$, train up to $10$ epochs, and select the best checkpoints based on performance on the validation set. Under default conditions, we use the ERNIE-base model and set the maximum input length to $512$ tokens. We conduct analytical experiments on hyperparameters such as model size and maximum length in Section~\ref{sec:hyperparameter}.

\subsection{Overall Performance}
\label{sec:backbone}
The results of benchmark models are shown in Table~\ref{table:model-selection}.
We can see that:
(1) All transformer-based methods achieve higher performance than classical methods.
This verifies the advantages of the Transformer-based architecture in natural language processing.
(2) Pre-trained models, especially the Ernie model, outperform other methods by a large margin.
This is not surprising because pre-training is widely regarded as the ability to incorporate external knowledge into model parameters.
(3) LLMs-based methods, since they are not sufficiently trained on the dialogue assessment task, still have a significant gap compared to the training-based approach. Further exploration is needed on how to better utilize LLMs for dialogue evaluation tasks.

Since the Ernie model achieves the best results on DiQAD, we use it as the backbone for detailed analysis in the next section.

\begin{table}[!t]\centering
\setlength\tabcolsep{3pt}
\resizebox{\columnwidth}{!}{
\begin{tabular}{@{}p{6.5em}p{3.5em}<{\centering}p{3.5em}<{\centering}p{3.5em}<{\centering}p{3.5em}<{\centering}p{3.5em}<{\centering}@{}}
\toprule
& Acc & UAR & Spearman & Pearson & F1 \\
\midrule
\multicolumn{4}{@{}l}{\emph{Digital domain}}\\
ERNIE & 63.89	& 38.40	& 50.14	& 50.08	& 63.90 \\
 - w/o Digital & 61.97	& 37.01	& 46.75	& 45.90	& 61.72 \\
\midrule
\multicolumn{4}{@{}l}{\emph{Emotion domain}}\\
ERNIE & 77.70	& 40.49	& 61.14	& 61.48	& 68.95 \\
 - w/o Emotion & 72.14	& 38.74	& 56.74	& 56.97	& 64.25\\
\midrule
\multicolumn{4}{@{}l}{\emph{Education domain}}\\
ERNIE & 69.82	& 40.55	& 52.13	& 54.78	& 68.06 \\
 - w/o Education & 64.26	& 37.21	& 45.13	& 44.76	& 62.54\\
\midrule
\multicolumn{4}{@{}l}{\emph{Legal domain}}\\
ERNIE & 72.63	& 41.00	& 56.73	& 55.71	& 69.27 \\
 - w/o Legal & 72.00	& 40.23	& 59.94	& 58.65	& 68.41\\
\midrule
\multicolumn{4}{@{}l}{\emph{Car domain}}\\
ERNIE & 67.71	& 38.56	& 47.93	& 47.03	& 64.49 \\
 - w/o Car & 66.67	& 38.84	& 48.92	& 48.06	& 64.32\\
\midrule
\multicolumn{4}{@{}l}{\emph{Finance domain}}\\
ERNIE & 68.35	& 38.96	& 53.61	& 52.77	& 65.80 \\
 - w/o Finance & 67.41	& 37.75	& 51.30	& 50.33	& 64.43\\
\bottomrule
\end{tabular}
}
\caption{Cross-domain evaluation results. Each group in the table represents the results of a specific domain.}\label{table:domain}
\end{table}

\subsection{Detailed Analysis}
\label{sec:detailed_analysis}
\paragraph{Cross-domain Generalizability}
Firstly, we analyze the cross-domain generalizability of the model trained on DiQAD.
We conduct comparison experiments by removing training samples from a particular domain during training and testing on that domain.
The experimental results are shown in Table~\ref{table:domain}.
For each domain, we also report Ernie's performance on it, where Ernie is trained with all training samples.

From Table~\ref{table:domain}, we can see that the model can still achieve comparable performance on domains not appearing in the training set.
For example, all accuracy scores drop by less than 6\% when the domain samples are removed from the training set.
We also find that the metric decline rates are related to the number of samples in that domain.
For example, the performance decrease of the model is more significant when a large domain (e.g., Emotion) is removed than a small domain (e.g., Finance).

\begin{figure}[!t]
 \centering
 \includegraphics[width=0.99\columnwidth]{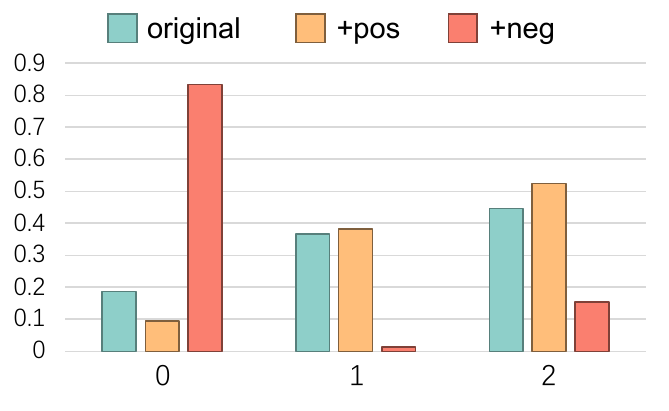}
 \caption{Comparison of predicted score distributions when adding different sentiment expressions.}
 \label{fig:sentiment}
 \vspace{-0.6cm}
\end{figure}

\paragraph{User Sentiment Influence}
To analyze the influence of user sentiment expression on dialogue quality assessment, we design two types of sentiment expressions and add them to the end of the original dialogues:
(i) Positive sentiment: We use the template ``\emph{question: ok thanks! response: you are welcome!}''\footnote{In Chinese: ``\emph{\begin{CJK}{UTF8}{gbsn}question: 好的多谢！response: 不谢！\end{CJK}}''}. We denote the modified model as \textbf{ERNIE+pos}.
(ii) Negative sentiment: We add the template ``\emph{question: I don't think so.}''\footnote{In Chinese: ``\emph{\begin{CJK}{UTF8}{gbsn}question: 不是这样的。\end{CJK}}''}. We denote the modified model as \textbf{ERNIE+neg}.
The experimental results are shown in Table~\ref{table:sentiment}.
We also demonstrate the prediction score changes when adding different sentiment expressions in Figure~\ref{fig:sentiment}.

From Figure~\ref{fig:sentiment} and Table~\ref{table:sentiment}, we can see that the model is more likely to predict a low-quality label when the dialogue contains negative user expressions and tends to give a higher quality if there is positive user feedback.
As a result, the performance of \textbf{ERNIE+pos} decreases slightly compared to the original Ernie (e.g., the accuracy score drops by around 3%);
and adding the negative expressions (i.e., \textbf{ERNIE+neg}) makes the accuracy score drastically decline by more than 38%.
These performance changes, especially on the negative sentiment, show that quality assessment in DiQAD is sensitive to sentiment expressions.
We believe that the sensitivity to user sentiment can help dialogue systems evolve in a direction that better satisfies the user.

% \paragraph{Robustness to non-semantic features}
% We summarize several non-semantic features and analyze the impact of these features on the model scores. 
% These features and the corresponding attack strategies are:
% (i) RepeatTurn: We repeat the last two dialogue turns to increase the number of dialogue turns.
% (ii) RepeatWord: We repeat the sentences of response to increase the average number of words.
% (iii) SafeResponse: When the last sentence of the dialogue is a question, we add a safe response.
% In Table we listed the results.
% %
% We can see that the model is generally robust to non-semantic features.
% Simply increasing dialogue turn or words in answers may not improve the score of the model.
% % \todo{}

\begin{table}[!t]\centering
\setlength\tabcolsep{1.5pt}
\resizebox{\columnwidth}{!}{
\begin{tabular}{p{5em}p{3.8em}<{\centering}p{3.8em}<{\centering}p{3.8em}<{\centering}p{3.8em}<{\centering}p{3.8em}<{\centering}}
\toprule
& Acc & UAR & Spearman & Pearson & F1 \\
\midrule
ERNIE & 69.94	& 40.96	& 57.39	& 57.25	& 68.46 \\
ERNIE+pos & 66.99	& 36.63	& 50.60	& 51.12	& 62.51 \\
ERNIE+neg & 31.33	& 24.55	& 27.07	& 28.05	& 27.04 \\
% \midrule
% ERNIE-RepeatTurn & 67.73	& 39.79	& 54.49	& 54.66	& 65.89 \\
% ERNIE-RepeatWord & 66.79	& 39.26	& 53.23	& 53.35	& 64.78 \\
% ERNIE-SafeResponse & 68.96	& 39.73	& 55.04	& 55.09	& 66.90 \\
\bottomrule
\end{tabular}
}
\caption{Comparison of evaluation results when adding different sentiment expressions.}\label{table:sentiment}
\end{table}
% \paragraph{Model size.} We use ERNIE of different sizes (e.g., mini, medium) and the results show that larger models achieve better performance. The detailed configurations of these models are available at Table~\ref{table:model-selection} in Appendix.

% \paragraph{Data size.} We random sample a portion of training data (e.g., 1\%, 5\%) and train the ERNIE-base model. We find that more data lead to better performance and the increase curve is basically in line with the logarithmic speed.

% \paragraph{Input length.} We change the maximize input token number and test the performance difference. 
% We find that increasing the maximum length can improve the effect when the length is less than $512$, and there is almost no improvement after the length exceeds $512$.

\subsection{Comparison with Other Reference-free Assessment Methods}
\label{sec:reference-free}

In this section, we compare the performances of other reference-free assessment methods on DiQAD.
There are usually two types of reference-free dialogue assessment methods:
unsupervised methods and supervised methods.
Unsupervised methods are usually learned without annotated dialogues, including:  USR~\citep{Mehri2020USRAU}, USL-H~\citep{Phy2020DeconstructTR}, BARTScore~\citep{Yuan2021BARTScoreEG}.
Supervised methods are usually learned upon annotated dialogues, including: BLEURT~\citep{Sellam2020BLEURTLR}, P-Tuning~\citep{Liu2021PTuningVP}, Prompted ERNIE (3-shot~\citep{Sun2021ERNIE3L}.

The evaluation metric is \emph{Spearman} and \emph{Pearson}, following previous studies~\cite{Mehri2020USRAU}.
The results are listed in Table~\ref{table:baselines}. We see that the unsupervised methods perform unfavorably. For example, USR and USL-H perform well on previous dialogue evaluation benchmarks but show low consistency with human assessment on DiQAD.
The main reason may be that in DiQAD, the user's questions are more difficult and the system needs to provide comprehensive responses to meet the user's needs, and simple metrics such as relevance and fluency are not well suited.
% probably because the proposed DiQAD task is more difficult, which makes existing unsupervised methods (e.g., use random negatives or rule-based samples) fail to model finer-grained dialogue quality in our task.
% (ii) The regression model (e.g., ERNIE regressor) performs slightly better on Spearman and Pearson compared to the classification model. However, we still mainly use the classification model because it can give a clear quality rating.

\begin{table}[!t]\centering
\resizebox{\columnwidth}{!}{
\begin{tabular}{@{}p{12.9em}p{3.5em}<{\centering}p{3.5em}<{\centering}@{}}
\toprule
& Spearman & Pearson \\
\midrule
\multicolumn{3}{@{}l}{\emph{Unsupervised methods}}\\
USR~\citep{Mehri2020USRAU} & \phantom{0}4.47 & 11.64\\
\quad - USR-MLM & \phantom{0}2.61 & \phantom{0}3.83\\
\quad - USR-DR & \phantom{0}3.83 & 11.61\\
USL-H~\citep{Phy2020DeconstructTR} & \phantom{0}4.60 & \phantom{0}7.91 \\
\quad - USL-VUP & \phantom{0}0.65 & ~-1.22\\
BARTScore~\citep{Yuan2021BARTScoreEG} & 24.73 & 32.77\\
\midrule
\multicolumn{3}{@{}l}{\emph{Supervised methods}}\\
BLEURT~\citep{Sellam2020BLEURTLR} & 48.37 & 49.96\\
 % - BLEURT-pre \\
% DEB~\citep{Sai2020ImprovingDE} \\
P-Tuning~\citep{Liu2021PTuningVP} & 53.03 & 53.02\\
Prompted ERNIE (3-shot) & 15.09 & 28.34\\
\quad - finetuning & 56.49 & 56.49\\
 % - Finetune\\
% ERNIE regressor & 57.98 & 57.96\\
\midrule
\textbf{ERNIE} & \textbf{57.39} & \textbf{57.25} \\
\bottomrule
\end{tabular}
}
\caption{Results of reference-free dialogue evaluation methods. The best results are shown in bold.}\label{table:baselines}
\vspace{-0.4cm}
\end{table}

\subsection{Hyperparameter Analysis}
\label{sec:hyperparameter}

\begin{figure}[!t]
 \centering
 \includegraphics[width=\columnwidth]{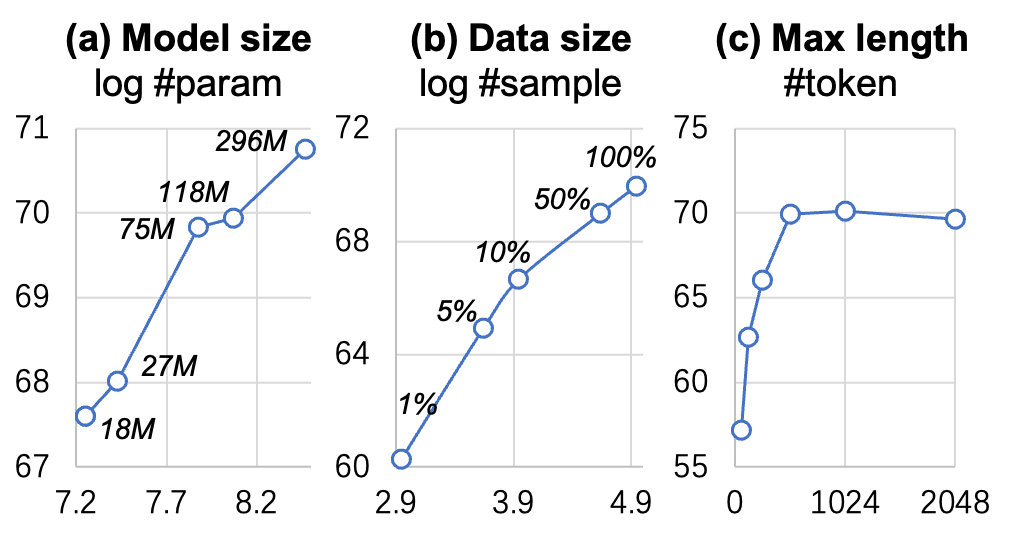}
 \caption{Hyperparameter analysis: (a) Model size, we demonstrate the Ernie performance with different numbers of parameters; (b) Data size, we show the effects of the training data size on the Ernie performance; (c) Max length, we compare the Ernie performances with different maximum input lengths.}
 \label{fig:hyperparameter}
 \vspace{-0.6cm}
\end{figure}

We conduct hyperparameter analysis, and the results are illustrated in Figure~\ref{fig:hyperparameter}. The aspect we analyzed and the findings are:
\begin{enumerate*}[label=(\alph*)]
    \item \textbf{Model size.} We use ERNIE of different sizes (e.g., 18M, 27M) and the results show that larger models achieve better performance.
    
    \item \textbf{Data size.} We random sample a portion of training data (e.g., 1\%, 5\%) and train the ERNIE-base model. We find that more data lead to better performance and the increase curve is basically in line with the logarithmic speed.

    \item \textbf{Input length.} We change the maximize input token number and test the performance difference. We find that increasing the maximum length can improve the effect when the length is less than $512$, and there is almost no improvement after the length exceeds $512$. This may be mainly because dialogue tokens in DiQAD are mostly less than 512.
\end{enumerate*}

%% file: 5-conclusion.tex
\section{Conclusion}
In this paper, we conduct a human-epistemic dialogue quality assessment on real user conversations.  
We release a large-scale dataset DiQAD for open-domain dialogue quality assessment, which contains around 100k dialogues and 0.8 million utterances in 6 domains.
We conduct several benchmark experiments on DiQAD to study the performances of classical, Transformer-based models and large language models on the dialogue quality assessment task. 
We hope the release of this dataset can inspire researchers in the community who lack such
real and large-scale datasets, to evaluate real user dialogues to guide further improvement for the language models' generation.
%In the future, we plan to expand the domain coverage of DiQAD by adding more kinds of dialogues e.g., multi-modal and multi-party dialogues.
%Besides, the current criteria of DiQAD do not involve factuality, which is usually regarded as important for dialogue system evolution.
%Therefore, another direction of our future work focuses on adding more quality dimensions (e.g., factuality).

%% file: 10_appendix.tex
\newpage
\appendix

\onecolumn

\section{Annotation Details}
\label{sec:appendix}

%\subsection{Annotation Details}

\subsection{Annotation Process}
We randomly sample the dialogues from WenYiWen platform that are licensed to be publicly available. We proceed to sample dialogues from six domains: emotion, digital, legal, education, car, and finance. 
After collect the dialogues from WenYiWen platform, we employee crowd-sourcing workers to annotate the quality labels for the dialogues (0, 1, 2).

The annotation process is an annotation-checking procedure, or repeated annotation-checking procedures (if the annotation accuracy of the first round of annotation does not meet the needs). The dialogues are divided into multiple batches for annotation, with each batch consisting of 2000 samples.

For each dialogue, the quality label have been agreed upon and confirmed by 2 crowd-source workers.
%In case of discrepancies, they would discuss or involve a annotation leader to determine the final rating. 
After annotate a batch of data, we have a professional quality controller to check the the label accuracy. (The roles and distribution of crowd-source workers can be found in section~\ref{sec:annotators}). Generally, a quality controller randomly samples 100 dialogues and check the label accuracy of this batch including overall annotation accuracy and the accuracy of each category (0, 1, 2). Our acceptance criterion is to achieve an overall accuracy of 92\% above, and the accuracy for each category no less than 90\% for each batch of data.

If the accuracy does not meet the thresholds, the entire batch of data is returned, and the existing labels are removed for re-annotation. This annotation and checking process is then repeated until the labeling accuracies surpass the thresholds. Based on the statistics of our annotation, approximately 31\% of batches require re-annotation, which means these data need two or more rounds of annotation.

It is worth noting that we continually annotate the dialogue quality data until now. And we sample 100k annotated dialogues for the open-domain dialogue quality assessment task for the community. We have applied the annotation in actual usage and were able to demonstrate real improvement when we use the data to train a dialogue quality model to determine the quality of real dialogues. The user satisfaction and user revisit ratio in WenYiWen platform have been improved 56\%, 23\%
relatively in the last year after deploying such dialogue quality evaluation.

\subsection{Annotation Instructions}
\label{subsec:annotation_instruction}

For each annotator, we provide instructions to enable them to perform the dialogue quality assessment task, as shown in Figure~\ref{fig:instruct_annotation}.

Each annotator take the form of a dialogue evaluation as shown in the following Figure~\ref{fig:interface_annotation}. This task includes to answer whether the information provided need to be verified through external knowledge in case of hallucinating problems, rate the quality score from -2, -1, 0, 1 to 2 and choose the reason why you rate the score. The -1 and -2 are rated when there are privacies and personal information contained in the responses and questions respectively. We use annotators to avoid the ethics.  We exclude the dialogues that score -2 and -1 for the consideration of ethics. This is the interface translated from Chinese, and the original Chinese interface is shown in Figure~\ref{fig:interface_ori_annotation}.

\begin{figure*}[!ht]
    \centering
    \includegraphics[width=1.0\linewidth]{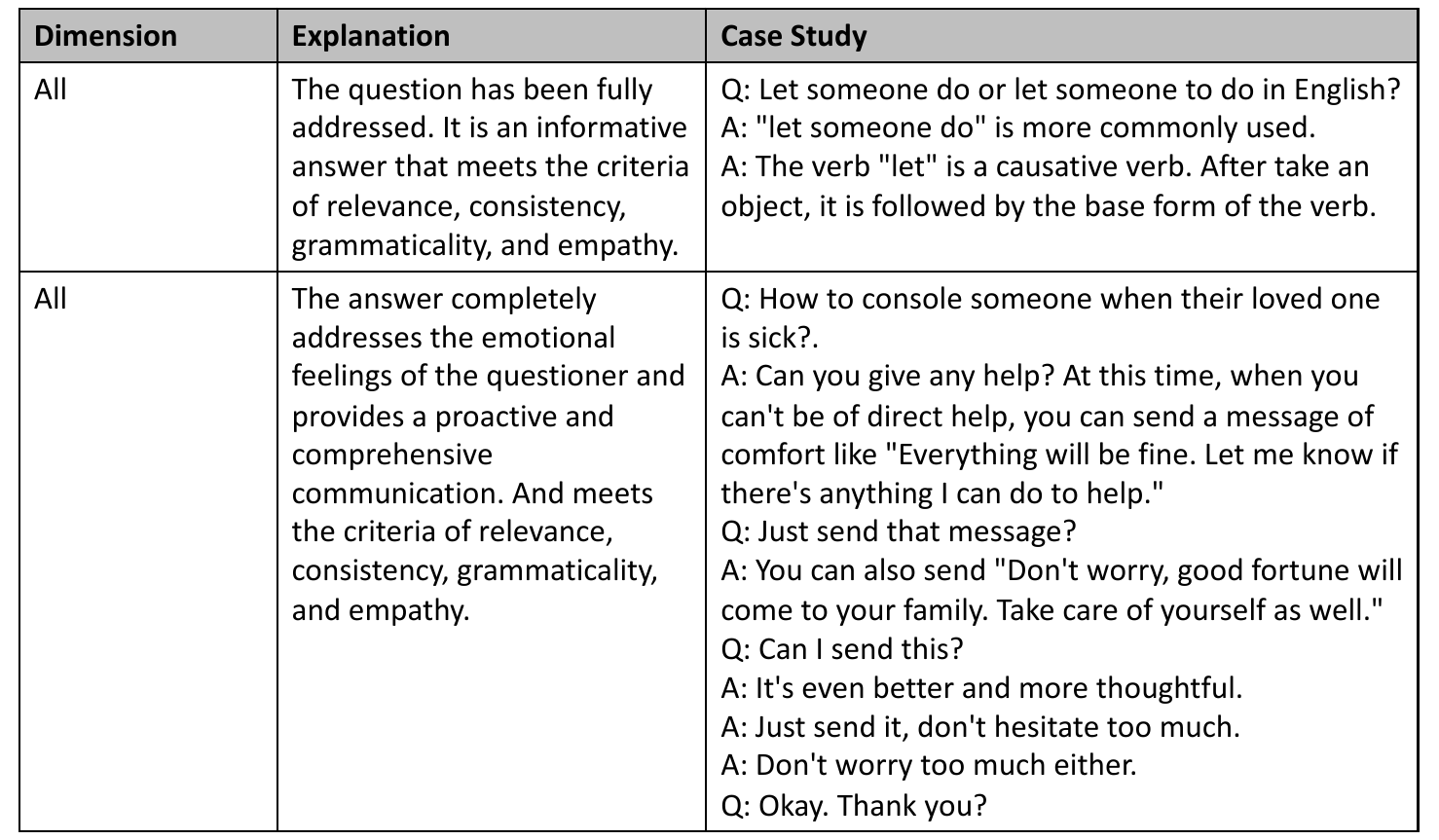}
    \caption{The detailed guidelines to rate 2 for the quality of a dialogue.}
    \label{fig:guide_annotation_2}
\end{figure*}

The detailed guidelines for scoring 0, 1, and 2 are shown in the following Figure~\ref{fig:guide_annotation_0}, Figure~\ref{fig:guide_annotation_1}, Figure~\ref{fig:guide_annotation_2} respectively.

\begin{figure*}[!ht]
    \centering
    \includegraphics[width=1.0\linewidth]{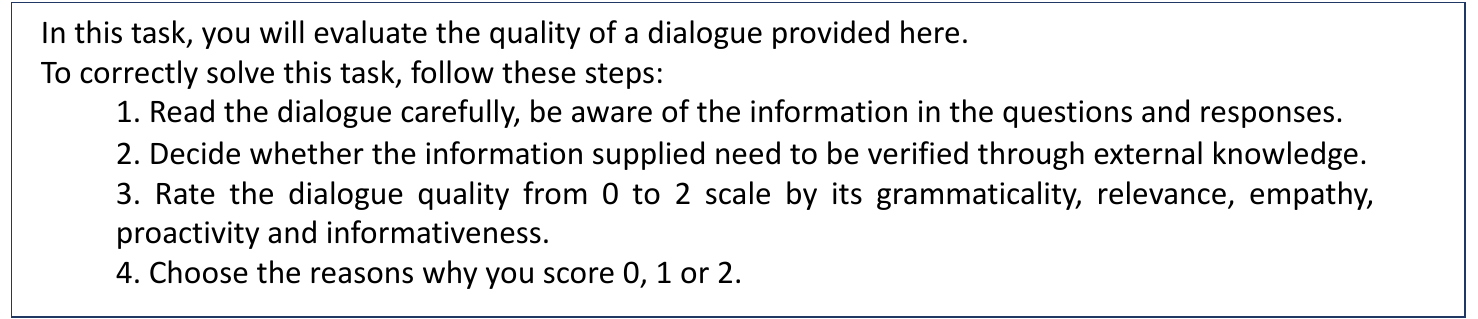}
    \caption{The task instruction used for dialogue quality evaluation.}
    \label{fig:instruct_annotation}
\end{figure*}

\begin{figure*}[!ht]
    \centering
    \includegraphics[width=1.0\linewidth]{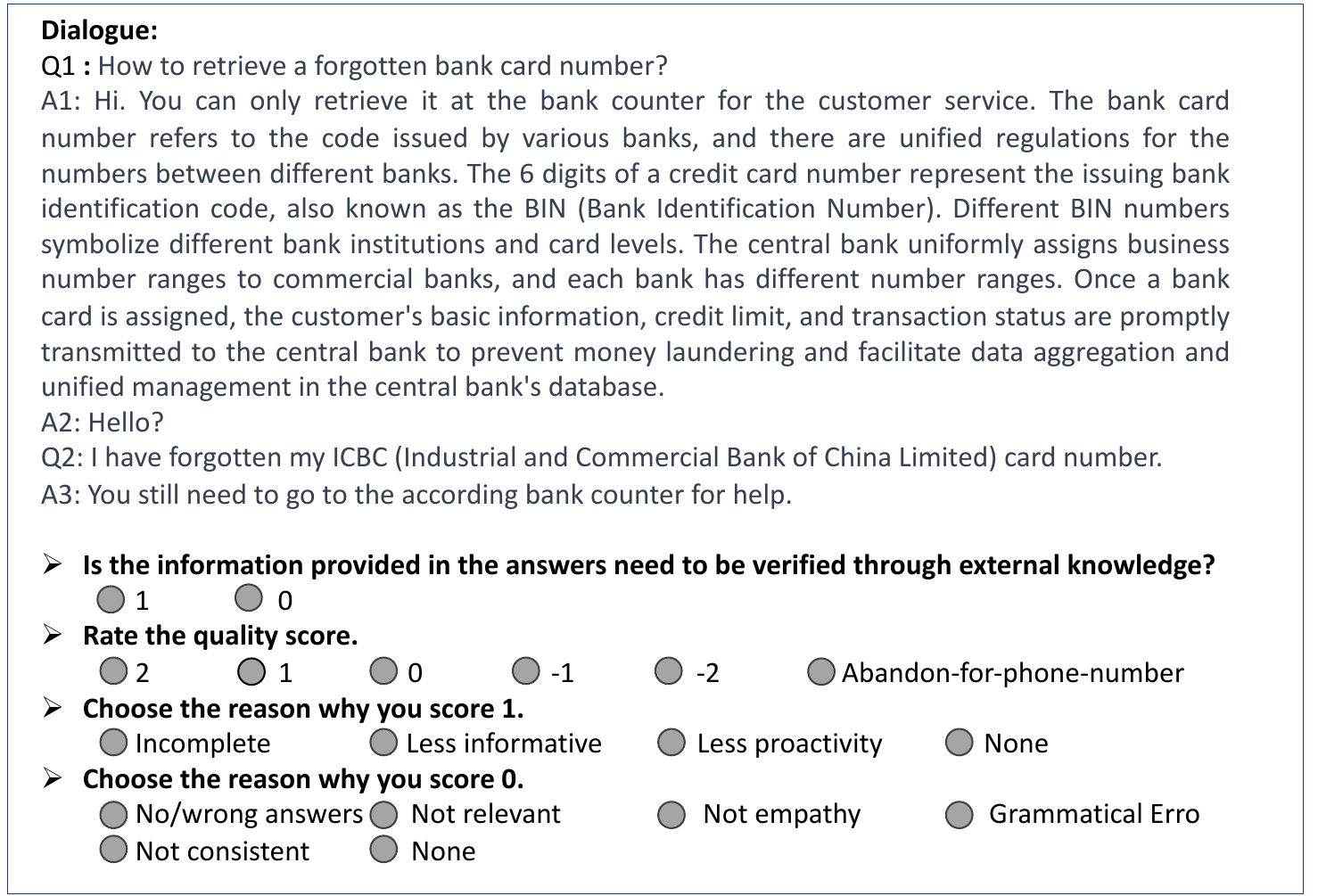}
    \caption{The interface for annotators to label the dialogue quality. This is a translated version from Chinese.}
    \label{fig:interface_annotation}
\end{figure*}

\begin{figure*}[!ht]
    \centering
    \includegraphics[width=1.0\linewidth]{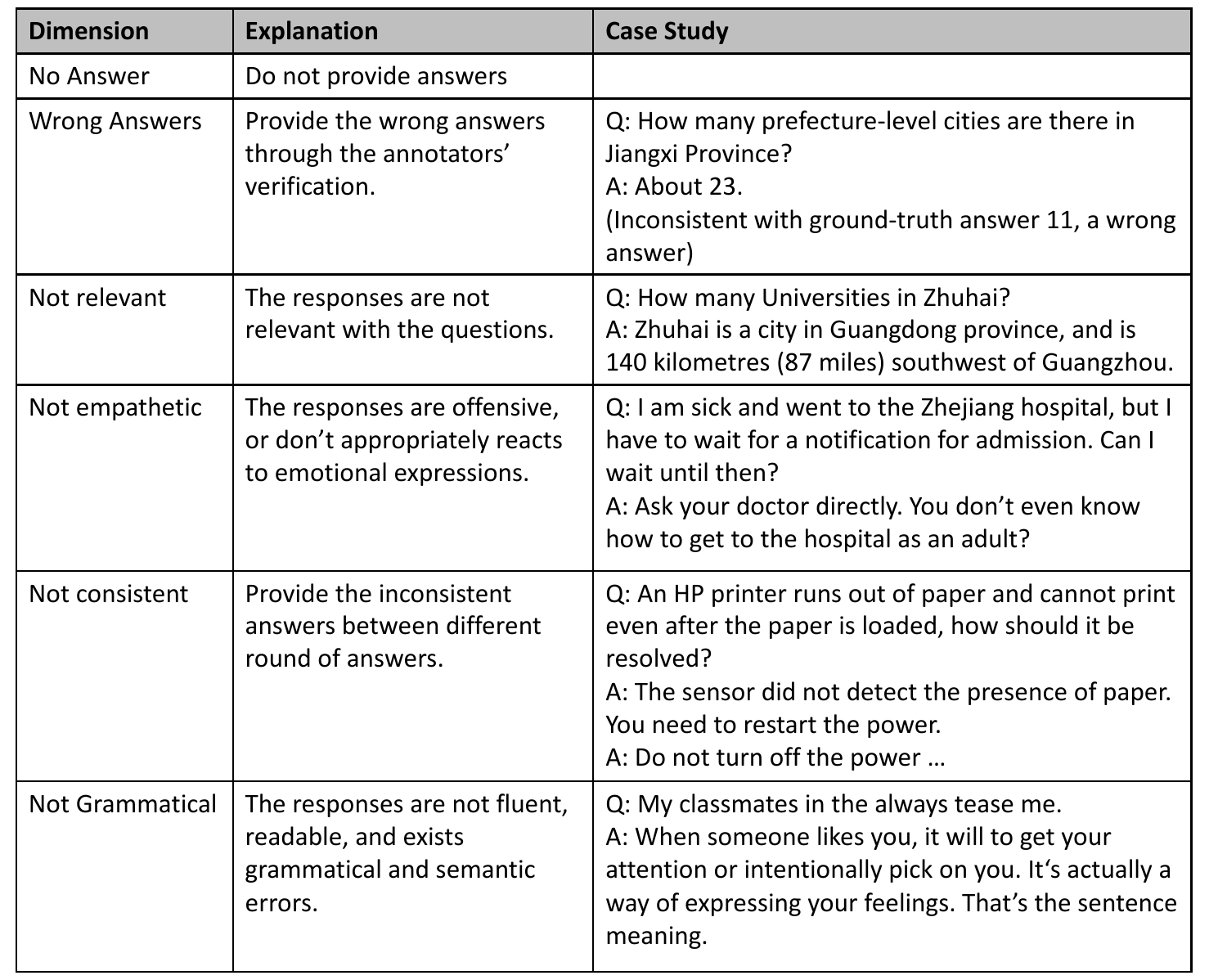}
    \caption{The detailed guidelines to rate 0 for the quality of a dialogue.}
    \label{fig:guide_annotation_0}
\end{figure*}

\begin{figure*}[!ht]
    \centering
    \includegraphics[width=1.0\linewidth]{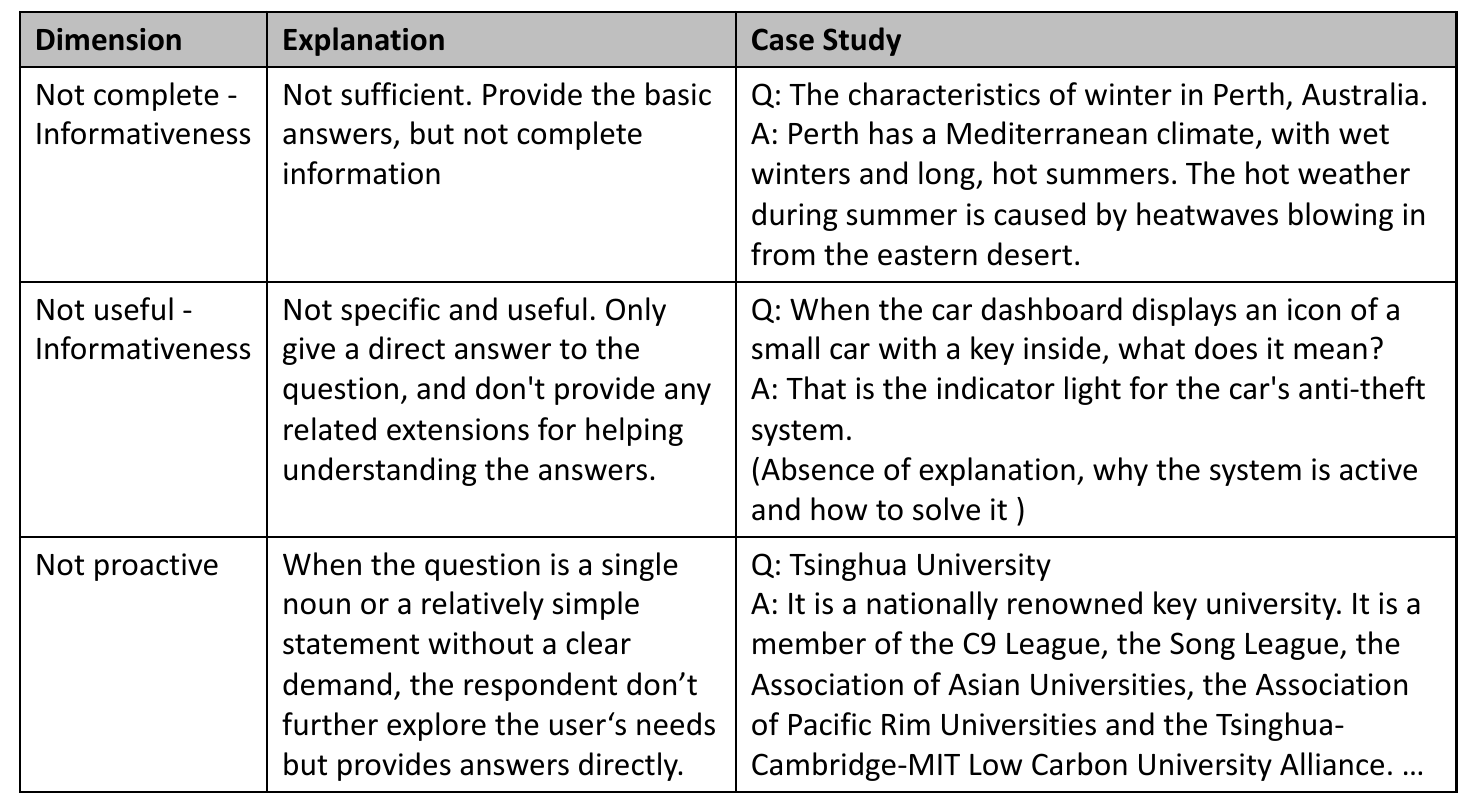}
    \caption{The detailed guidelines to rate 1 for the quality of a dialogue.}
    \label{fig:guide_annotation_1}
\end{figure*}

\begin{figure*}[!ht]
    \centering
    \includegraphics[width=1.0\linewidth]{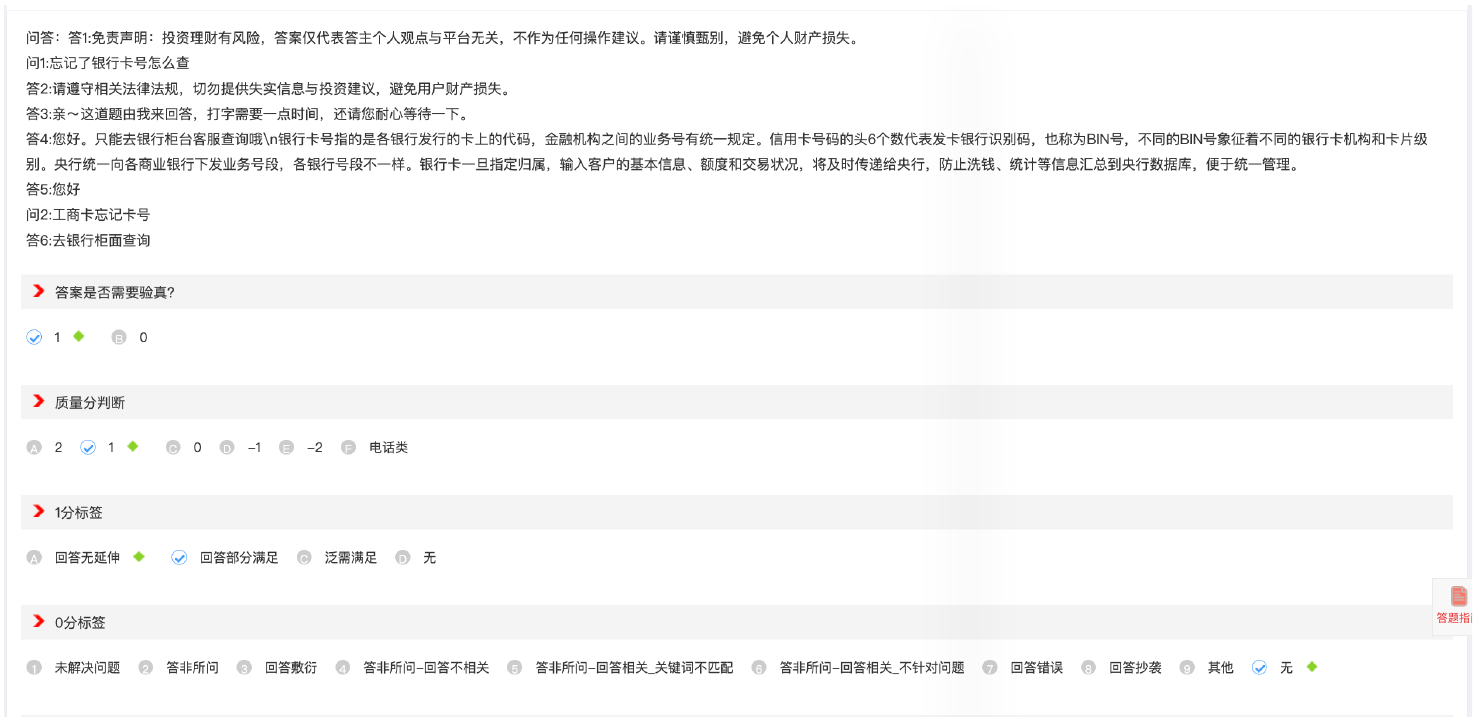}
    \caption{The interface for annotators to label the dialogue quality in Chinese.}
    \label{fig:interface_ori_annotation}
\end{figure*}

\subsection{Annotators}
\label{sec:annotators}

The annotators are full-time crowdsourcing annotators who continuously annotate the dialogues to serve the commercial WenYiWen platform. We sample 100k dialogues to be published to the community for the dialogue quality evaluation task.

We maintain an annotation team of approximately 26 employees, including 2 trainers, 4 quality-controllers, and 20 annotators. All of them have at least a colledge degree. The 20 annotators carry out specific annotations, while the quality-controllers randomly review 100 annotations for each batch data after it has been annotated, to assess the annotatation accuracy against demands. Trainers are responsible for collecting cases of inconsistent annotations, discussing and formulating the annotation metrics with us and training the reviewers and annotators using the annotation criteria.

\section{Baseline Details}

Unsupervised methods are usually learned without annotated dialogues, which include:
\begin{itemize}
    \item \textbf{USR}~\citep{Mehri2020USRAU}, which is a holistic score for quality evaluation using the RoBERTa as the backbone. USR is composed of two sub-scores: USR-MLM and USR-DR, where USR-MLM is calculated using the log-likelihood values of responses, and USR-DR finetunes the model with randomly sampled negatives.
    \item \textbf{USL-H}~\citep{Phy2020DeconstructTR} is composed of VUP, NUP, and MLM, in which NUP and MLM are basically the same as USR-DR and USR-MLM, respectively, and VUP finetunes the model with synthetic negatives.
    \item \textbf{BARTScore}~\citep{Yuan2021BARTScoreEG}, which uses the log-probability of BART generating answers conditioned on questions as the score.
\end{itemize}

Supervised methods are usually learned upon annotated dialogues, including:
\begin{itemize}
    \item \textbf{BLEURT}~\citep{Sellam2020BLEURTLR}, in which a BERT-large model is pre-trained on synthetic data and fine-tuned with regression loss; 
    \item \textbf{P-Tuning}~\citep{Liu2021PTuningVP}, which optimizes the parameters of the prompt vector and freezes other parameters of ERNIE;
    \item \textbf{Prompted ERNIE (3-shot)}~\citep{Sun2021ERNIE3L}, which uses 3 annotated dialogues as in-context examples to prompt an ERNIE model; we also test supervised finetuning setting by finetuning the parameters of the prompted model on the training set.
\end{itemize}

\section{Prompts}\label{sec:prompt}

\newtcolorbox{promptbox}[1]{colback=Gray!5!white,colframe=Gray!95!black,fonttitle=\bfseries,title=#1}

\begin{promptbox}{Chinese prompt}
\begin{CJK*}{UTF8}{gbsn}
\textbf{USER}

我会给出来自user和assistant的对话。
对话以json格式保存，每项包括了其角色role（user或assistant）和消息内容content。
请根据assistant的表现对其回复质量打分，标准如下：

- 0分，低质量

- 1分，中等质量

- 2分，高质量

注意你只需要输出一个分数代表回复的质量，不需要做任何解释。\\

\textbf{ASSISTANT}

好的，请提供对话，我将给出0到2的分数。\\

\textbf{USER}

[{'role': 'user', 'content': '不超过十字的新闻。快！快！'}, {'role': 'assistant', 'content': '再短的新闻也很难低于40个字。 不超过十字的就是新闻标题，如下 香港机场航班大致正常，少量取消部分延误 天津一中学生遭班主任体罚横纹肌溶解 金融委全面加大对实体经济的支持力度 国庆假期火车票明天开抢'}]\\

\textbf{ASSISTANT}

分数：0\\

\textbf{USER}

[{'role': 'user', 'content': '平安保险生效7天2万还能退多少'}, {'role': 'assistant', 'content': '您好，很感谢您的提问，可以全额退款，因为生效7天还属于犹豫期，犹豫期退保是全额退的。希望我的回答能够帮助到您，祝您生活愉快！'}]\\

\textbf{ASSISTANT}

分数：1\\

\textbf{USER}

[{'role': 'user', 'content': '那去哪里可以查的到自己有没有被起诉呢 '}, {'role': 'assistant', 'content': '通常去调取信息的话，您可以选择去公安局或是法院，提供您和您要查询的案件或当事人的身份关系，另外还有您的身份证以及户口本，还有就是说明您要查询的原因（这个法院那边会让您填表或是给您直接进行登记的）。您可以带好这些材料，法院那边都是可以帮您查的。'}]\\

\textbf{ASSISTANT}

分数：2\\

\textbf{USER}

\texttt{\{\{dialogue\}\}}

\end{CJK*}
\end{promptbox}

\newtcolorbox{enbox}[1]{colback=Gray!5!white,colframe=Gray!95!black,fonttitle=\bfseries,title=#1}

\begin{enbox}{English prompt (translated version)}
\textbf{USER}

I will provide you a dialogue between user and assistant. The dialogue is saved in json format, in which each item include the role (user or assistant) and the message content.
please rate the assistant performance based on the response quality, and the criterion is as follow:

- Score 0, low quality.

- Score 1, moderate quality.

- Score 2, high quality.

Note that you only need to output the number of the score, without any explanation\\

\textbf{ASSISTANT}

OK, please provide the dialogue and I will rate it in scale of 0-2.\\

\textbf{USER}

[{'role': 'user', 'content': 'News that does not exceed the cross. Hurry up! Hurry up!'}, {'role': 'assistant', 'content': 'Even the shortest news can hardly be less than 40 words. No more than the cross is the Headline, as shown below, flights at the Hong Kong airport are generally normal, a small number of cancellations and partial delays, a middle school student in Gamma Cygni was physically punished by the head teacher, and the Striated muscle tissue was dissolved. The Financial Committee fully increased its support for the real economy. Train tickets will be robbed tomorrow during the National Day holiday'}]\\

\textbf{ASSISTANT}

Score: 0\\

\textbf{USER}

[{'role': 'user', 'content': 'How much can I refund 20000 yuan after Ping An Insurance takes effect for 7 days'}, {'role': 'assistant', 'content': 'Hello, thank you very much for your question. We can offer a full refund as the 7-day validity period is still a hesitation period, and refunds during the hesitation period are full refunds. I hope my answer can help you, and I wish you a happy life!'}]\\

\textbf{ASSISTANT}

Score: 1\\

\textbf{USER}

[{'role': 'user', 'content': 'Where can I find out if I have been prosecuted'}, {'role': 'assistant', 'content': 'If you usually need to retrieve information, you can choose to go to the public security bureau or court to provide your identity and relationship with the case or party you want to inquire about, as well as your ID card and household registration book, and explain the reason for your inquiry (the court will ask you to fill out a form or directly register it for you). You can bring these materials with you, and the court can help you investigate them.'}]\\

\textbf{ASSISTANT}

Score: 2\\

\textbf{USER}

\texttt{\{\{dialogue\}\}}

\end{enbox}